\documentclass[10pt,twocolumn,letterpaper]{article}
\usepackage{iccv}

\usepackage{times}
\usepackage{url} 
\usepackage{epsfig}
\usepackage{graphicx}
\usepackage{amsmath}
\usepackage{amssymb}
\usepackage{paralist}
\usepackage{caption}
\usepackage{xcolor}
\usepackage[pagebackref=true,breaklinks=true,colorlinks,bookmarks=false]{hyperref}

\iccvfinalcopy
\def\iccvPaperID{3433} % *** Enter the ICCV Paper ID here

\ificcvfinal\fi

\begin{document}

%%%%%%%%% TITLE
\title{The ThreeDWorld Transport Challenge: A Visually Guided \\ Task-and-Motion Planning Benchmark for Physically Realistic Embodied AI 
}
\author{Chuang Gan$^{1}$
\quad
Siyuan Zhou$^{2}$
\quad
Jeremy Schwartz $^2$ 
\quad
Seth Alter$^2$ 
\quad
Abhishek Bhandwaldar$^1$ 

\\

Dan Gutfreund$^1$ 
\quad

Daniel L.K. Yamins$^3$ \quad
James J. DiCarlo$^2$ \quad
Josh McDermott$^2$ \\
Antonio Torralba$^2$
\quad
Joshua B. Tenenbaum$^2$\\\\
$^1$ MIT-IBM Watson AI Lab \quad  $^2$ MIT \quad
$^3$ Stanford University \\
\url{http://tdw-transport.csail.mit.edu}  
}

% \begin{figure*}[t]
%     \centering
%     \includegraphics[width=0.9\linewidth]{Figure/teaser.pdf}
%     \caption{An overview of TDW-Transport Challenge.  In this example task, the agent must transport two objects from the floor in one room and place them on the bed in the bedroom. The agent can first pick up the container, put two objects into it, and then transport them to the target location.}
%     \label{fig:my-challenge-figure}
% \end{figure*}

% \twocolumn[{%
% \renewcommand\twocolumn[1][]{#1}%
% %\vspace{-1em}
% \maketitle
% %\begin{center}
%     \centering
%     \includegraphics[width=0.95]{Figure/teaser.pdf}
%     \captionof{figure}{An overview of TDW-Transport Challenge.  In this example task, the agent must transport two objects from the floor in one room and place them on the bed in the bedroom. The agent can first pick up the container, put two objects into it, and then transport them to the target location. 
%     }
%     \label{fig:my-challenge-figure}

% }]

\twocolumn[{%
\renewcommand\twocolumn[1][]{#1}%
%\vspace{-1em}
\maketitle
%\begin{center}
    \centering
    \vspace{-4mm}
    \includegraphics[width=0.85\linewidth]{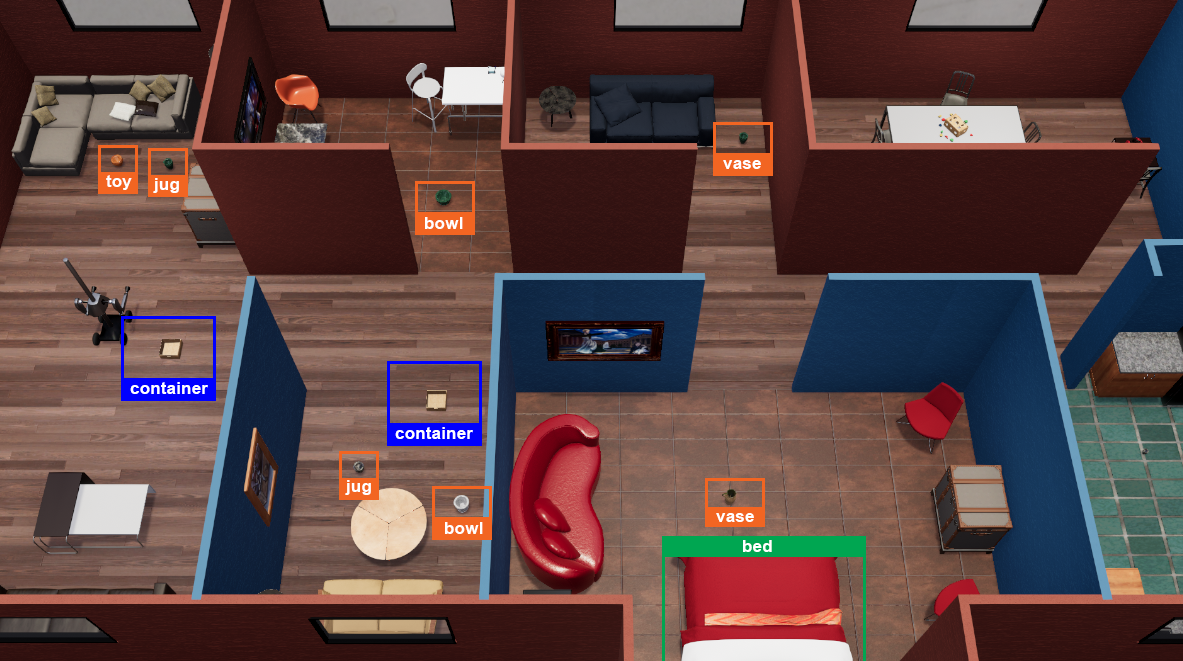}
    \captionof{figure}{An overview of the ThreeDWorld Transport Challenge.  In this example task, the agent must transport objects scattered across multiple rooms and place them on the bed (marked with a green bounding box) in the bedroom. The agent can first pick up a container, put objects into it, and then transport them to the goal location.  
    }
    \label{fig:my-challenge-figure}
\vspace{3.5mm}
}]

%%%%%%%%% ABSTRACT
\begin{abstract}
We introduce a visually-guided and physics-driven task-and-motion planning benchmark, which we call the ThreeDWorld Transport Challenge. In this challenge, an embodied agent equipped with two 9-DOF articulated arms is spawned randomly in a simulated physical home environment. The agent is required to find a small set of objects scattered around the house, pick them up, and transport them to a desired final location.  We also position containers around the house that can be used as tools to assist with transporting objects efficiently. To complete the task, an embodied agent must plan a sequence of actions to change the state of a large number of objects in the face of realistic physical constraints. We build this benchmark challenge using the ThreeDWorld simulation: a virtual 3D environment where all objects respond to physics, and where can be controlled using a fully physics-driven navigation and interaction API. We evaluate several existing agents on this benchmark. Experimental results suggest that: 1) a pure RL model struggles on this challenge;  2)  hierarchical planning-based agents can transport some objects but still far from solving this task.  We anticipate that this benchmark will empower researchers to develop more intelligent physics-driven robots for the physical world.
\end{abstract}
%%%%%%%%% BODY TEXT

\section{Introduction}

Household robots able to sense and act in the physical world remain an important goal of the computer vision and robotics communities. Because directly training models with real robots is expensive and involves safety risks, there has been a trend toward incorporating simulators to train and evaluate AI algorithms. In recent years, the development of 3D virtual environments~\cite{kolve2017ai2,xia2018gibson,savva2019habitat,xiang2020sapien,gan2020threedworld} that can simulate photo-realistic scenes has served as the major driving force for the progress of vision-based robot navigation~\cite{zhu2017target,gupta2017cognitive,chaplot2020learning,anderson2018vision,ramakrishnan2020occupancy,shridhar2020alfred,chen2019learning}.

Most tasks defined thus far in these virtual environments primarily focus on visual navigation in high-quality synthetic scenes~\cite{song2017semantic} and real-world RGB-D scans~\cite{chang2017matterport3d,xia2018gibson}, and thus pay little or no attention to physical interaction. Since the ultimate goal of Embodied AI is to develop systems that can perceive and act in physical environments, it has become increasingly apparent that physical interaction is a necessary component for the training of home assistant robots.  Recent work has aimed to improve sim-to-real transfer in the physical world by developing 3D virtual environments with both photo-realistic rendering and high-fidelity physics simulations (\eg iGibson~\cite{xia2020interactive},   TDW~\cite{gan2020threedworld} and SAPIEN ~\cite{xiang2020sapien}). iGibson~\cite{xia2020interactive,xia2020relmogen} made initial progress in this area, introducing an interactive navigation task with partial observations in a 3D cluttered environment. An agent with a mobile manipulator was encouraged to take interactive action (\eg pushing obstacles away and opening a door) to navigate towards the goal position. But the task itself was still a relatively simple navigation task, and did not explicitly require agents to change object states.

In this paper, we propose a new embodied AI challenge: an agent must take actions to move and change the state of a large number of objects to fulfill a complex goal in a photo- and physically-realistic virtual environment.  In our challenge, an embodied agent equipped with two 9-DOF articulated arms is spawned randomly in a physically-realistic virtual house environment. The agent is required to explore the house, searching for a small set of objects scattered in different rooms, and transporting them to a desired final location, as shown in Fig.\ref{fig:my-challenge-figure}. 
We also position various containers around the house; the agent can find these containers and place objects into them. Without using a container as a tool, the agent can only transport up to two objects at a time. However, using a container, the agent can collect several objects and transport them together. 

We build our challenge using ThreeDWorld (TDW)~\cite{gan2020threedworld}, an interactive physical simulation platform.  We first create a TDW-based house dataset of multi-room environments filled with objects that respond to physics.   We further develop a fully physics-driven high-level navigation and interaction API that we can use to train AI agents to physically interact with the virtual world. The embodied agent needs to pick up, move and drop objects using physics to drive a set of joints; this is fundamentally different from the approach taken by AI2-THOR~\cite{kolve2017ai2} and VirtualHome~\cite{puig2018virtualhome}, where objects close to the agent simply ``attach" to it or are animated to move into position, with no physical interaction at all.  We believe that such a long-horizon task in the physically realistic environment poses several challenges for embodied agents beyond semantic exploration of unknown environments, including:
\begin{compactitem}
\item \textbf{Synergy between navigation and interaction. } The agent cannot move to grasp an object if this object is not in the egocentric view, or if the direct path to it is obstructed (\eg by a table). 

\item \textbf{Physics-aware Interaction.} Grasping might fail if the agent's arm cannot reach an object.

\item \textbf{Physics-aware navigation.} Collision with obstacles might cause objects to be dropped and significantly impede the transport efficiency.

\item \textbf{Reasoning about tool usage.} While the containers help the agent transport more than two items, it also takes some time to find them. The agent thus has to reason about a case-by-case optimal plan. 
\end{compactitem}

We evaluate several existing agents on this benchmark. Experimental results suggest that existing state-of-the-art models for embodied intelligence all struggle to complete this task efficiently. We believe models that perform well on our transport challenge will enable more intelligent robots that can function in the real physical world. Our contributions are summarized as follow:
\begin{compactitem}
  \item We introduce a new embodied AI task that aims to measure AI agents' ability to change the states of multiple objects to accomplish a complex task in a photo- and physically-realistic virtual environment.
    
%  \item We have created a ThreeDWorld-House dataset that contains 50 object categories that respond to physics. 
  
  \item We have developed a fully physics-driven high-level command API that enables agents to execute interactive actions in this simulated physical world. 
   
  \item We evaluate several agents and find that a pure RL  model struggles to succeed at the challenge task. The hierarchical planning-based agent achieves better performance but still far from solving this task. 
    
\end{compactitem}

\section{Related Work}

Our challenge builds on prior work on 3D interactive environments, embodied intelligence and task-and-motion planning.

\subsection{3D Interactive Environments}
Recently, the development of 3D interactive environments, including AI2-THOR\cite{kolve2017ai2}, HoME~\cite{wu2018building}, VirtualHome~\cite{puig2018virtualhome}, Habitat~\cite{savva2019habitat},  Gibson~\cite{xia2018gibson}, Deepmind Lab~\cite{beattie2016deepmind}, RoboTHOR~\cite{deitke2020robothor}, SAPIEN~\cite{xiang2020sapien}, and TDW ~\cite{gan2020threedworld} have facilitated novel algorithms in visual navigation~\cite{zhu2017target,savva2019habitat,ramakrishnan2020occupancy}, visual-language navigation~\cite{anderson2018vision,wang2018look,shridhar2020alfred}, embodied question answering~\cite{das2018embodied,gordon2018iqa} and other tasks. 

There are several stand-alone physics engines widely used in the robotics community, including PyBullet~\cite{coumans2016pybullet},  MuJuCo~\cite{todorov2012mujoco}, and V-REP~\cite{rohmer2013v}. Many robotic manipulation and physical reasoning challenges are also built on top of these engines ( \eg RL-Benchmark~\cite{james2020rlbench}, Meta-World \cite{yu2020meta} and CLEVRER \cite{yi2019clevrer,chendlc21}). However, these platforms cannot render photo-realistic images, limiting the visual perception abilities of systems trained therein. Most recently, physics-based simulators with realistic images and physics have been introduced~\cite{xiang2020sapien,xia2020interactive,gan2020threedworld}, aiming to reduce the sim-to-real gaps seen when deploying trained systems in the physical world.  However, we still lack challenging embodied AI benchmarks with a clear task and evaluation metric that require an agent to move and change multiple object states for a long-horizon task.  We hope to fill this gap with the ThreeDWorld Transport Challenge. Our challenge involves complex physical scene understanding and a long-horizon task and can serve as a benchmark for running and testing embodied agents' task-and-motion planning abilities in 3D simulated physical home environments.

\subsection{Embodied Intelligence}
The training of embodied agents in 3D virtual environments has generated increasing interest within the computer vision and robotics communities. Popular tasks include point-goal navigation~\cite{savva2019habitat}, audio-visual navigation~\cite{gan2019look,chen2019audio}, vision-language navigation~\cite{anderson2018vision}, and semantic-goal navigation~\cite{chaplot2020object}. However, a majority of these navigation tasks are collision-free and do not involve much physical interaction.   

Inspired by early work on Navigation Among Movable Objects (NAMO)~\cite{stilman2008planning,stilman2007manipulation,van2009path,van2009path}, iGibson~\cite{xia2020interactive} introduced an interactive navigation task in a cluttered 3D environment. In~\cite{xia2020interactive}, an agent is encouraged to push movable objects with non-prehensile manipulators to clear a path to aid navigation. In~\cite{xia2020relmogen}, more complicated interactive navigation tasks (\eg open a door) and a few mobile manipulation tasks (\eg kitchen rearrangement) are developed. However, in both cases, the tasks have a shorter horizon and involve limited physical constraints compared with ours.  We aimed to use a more advanced physical virtual environment simulator to define a new embodied AI task requiring an agent to change the states of multiple objects under realistic physical constraints. Perhaps the closest challenge task to that proposed here is a recently introduced Animal AI Olympic challenge~\cite{beyret2019animal}. This challenge aims to test AI reasoning and planning abilities in a Unity game engine. However, the embodied agent is restricted to navigation actions, and does not contain interactive actions that are important for building household robots.  Our new benchmark goes beyond these existing challenge datasets by requiring the agents to change multiple object states in a rich and realistic physical world. Hierarchical reasoning and planning abilities in the face of realistic physics constraints are necessary to succeed at our task.

Current state-of-the-art methods on embodied tasks mainly fall into two broad categories:  end-to-end training of neural policies using RL~\cite{zhu2017target, mirowski2016learning} or hierarchical RL~\cite{kaufmann2019beauty,bansal2020combining} and map building for path planning~\cite{gupta2017cognitive,savinov2018semi,chaplot2020learning}. We believe that the proposed ThreeDWorld transport challenge could provide new opportunities to develop novel algorithms that combine visual perception, reasoning and hierarchical planning to solve more challenging tasks in the physical world.

% \subsection{Benchmarks for Physics Scene Understanding} Our work is also closely related to physical scene understanding~\cite{wu2015galileo,battaglia2013simulation,agrawal2016learning,
% wu2017learning,innamorati2019neural,ehsani2020use}. Physical dynamics is of great importance for enabling deep-learning based approaches to model-based planning and control applications~\cite{Lerer2016Learning,Mottaghi2016What,Fragkiadaki2016Learning,Battaglia2016Interaction,Chang2017compositional,Finn2016Unsupervised,Shao2014Imagining,fire2016learning,pearl2009causality,YeTian_physicsECCV2018}.
% Recently, several datasets have also been developed for physics reasoning.  CLEVRER~\cite{yi2019clevrer} studied visual dynamic reasoning in collision events in a video question answering form.  Intphys\cite{riochet2018intphys} introduced a synthetic dataset for visual intuitive physics reasoning.  COPHY~\cite{baradel2019cophy} studied counterfactual physical dynamics predictions in 3 different scenarios. Phyre~\cite{bakhtin2019phyre} proposed several 2D puzzles to measure agents' ability to use tools.

\subsection{Task and Motion Planning}
Our work falls in the domain of task and motion planning (TAMP)~\cite{kaelbling2011hierarchical,kaelbling2013integrated,garrett2015ffrob,wang2018active}, recently reviewed in~\cite{garrett2020integrated}. The goal of TAMP is to operate a robot in environments containing a large number of objects, taking actions to move and change the state of the objects in order to fulfill a goal. A task planner is responsible for reasoning over the large set of states of the environment, and a motion planner computes a path to accomplish the task. TAMP has demonstrated its power in many mobile manipulation tasks~\cite{wolfe2010combined,srivastava2014combined}. In our work, we hope to create a visually guided TAMP-like challenge that requires robots to carry out a complex and long-horizon task involving physical interaction in a cluttered 3D environment. Rearrangement of objects in an environment has recently been proposed as a potential test bed for embodied AI~\cite{batra2020rearrangement}. Our work provides an example rearrangement challenge with a visually guided task-and-motion planning benchmark in a photo- and physically-realistic virtual environment.

\begin{figure*}[t]
    \centering
   \includegraphics[width=1\linewidth]{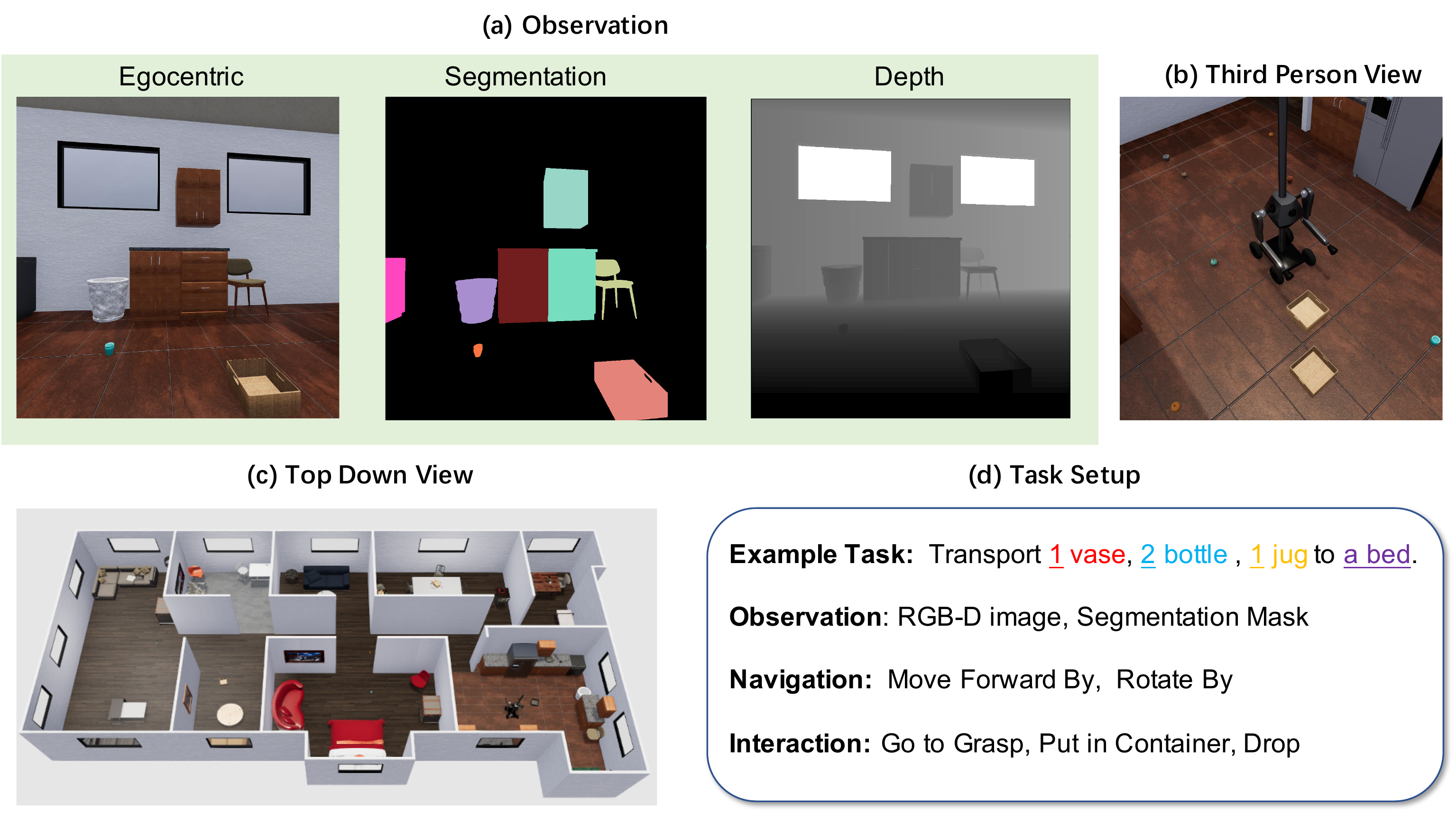}
    \caption{The details of ThreeDWorld Transport Challenge. (a) The observation state includes first-person view RGB image, Depth image, and semantic segmentation mask; (b) and (c) are Third-person view, and top-down view of the environment respectively; (d) Outline of the task and action space. }
    \label{fig:config}
\end{figure*}
\section{ThreeDWorld Transport Challenge}

The ThreeDWorld Transport challenge aims to assess an AI agent's reasoning and planning abilities in a physical realistic environment.  We build this challenge on top of the TDW platform~\cite{gan2020threedworld}, which is a general-purpose virtual world simulation platform supporting both near-photo-realistic image rendering, physically-based sound rendering~\cite{traer2019impacts}, and realistic physical interactions between objects and agents. 

TDW enables the training of embodied agents to perform tasks in a simulated 3D physical world. To support such training, we created 15 physically-simulated houses filled with objects that respond to physics. We also developed a high-level fully physics-driven navigation and interaction API. An agent equipped with an RGB-D camera and an oracle visual perception model can navigate through the virtual physical world to transport objects. Since the simulated actions and environment are fully physics-based, they pose additional challenges compared to previous non-physics~\cite{savva2019habitat,kolve2017ai2,puig2018virtualhome} or partial-physics~\cite{xia2020interactive} virtual environments. For instance, interactive action only succeeds if the target is physically reachable (\ie close by and not obstructed).  The agent can not successfully go to grasp an object if this object is not in the egocentric view, or if the direct path to it is obstructed (\eg by a table).  And physical collisions with objects in the house might also significantly impede transport progress. Therefore, the agent must learn to leverage visual signals to synchronize navigation and manipulation under these physical constraints.

\subsection{Problem Formulation}
 
As shown in Figure~\ref{fig:config}, an embodied agent is required to transport a set of predefined objects to a target location. There are three types of objects that the agent needs to pay attention to: 
\begin{compactitem}
    \item \textbf{Target objects:} objects the agent needs to grasp and transport, placed at various locations in the house. 

  \item \textbf{Containers:}  objects that can be used as tools to assist with the transport of target objects, also located in various rooms.
  
  \item \textbf{Goal position:} a goal area is defined around one unique object (e.g, a bed) in  the house.
  \end{compactitem}

The objective of this task is to transport as many objects as possible, given a limited interaction budget. The robot agent can only transport two objects with their arms, but it can transport more with the container. In each task instance, the target object positions and the object list for transport are varied, so the agent must learn how to explore the home environment to find target objects, containers, and goal positions. They can then plan a sequence of actions to transport the target objects to their desired location. Since both the objects in the room and high-level actions are responsive to physics, the robot agent needs to learn to use vision to decide when to go to grasp objects that can be successfully reached while avoiding collisions (\eg objects and/or the container might fall if the agent collides with obstacles) under the realistic physic constraints.

\subsection{Scenes}

We have created a TDW-House dataset to support this challenge. As shown in Figure \ref{fig:config} (c), each house environment contains between 6 and 8 interconnected rooms such as bedrooms, living rooms, and kitchens. These rooms are fully populated with furniture and other items (see ``Objects" below). The agent's view might be blocked by obstacles such as walls, so it needs to explore the house in order to predict a full occupancy map of the scenes.

The dataset is modular in its design, comprising several physical floor plan geometries with wall and floor material variations (e.g. parquet flooring, ceramic tile, stucco, carpet etc.) and various furniture and prop layouts (tables, chairs, cabinets etc.), for a total of 15 separate environments. Furniture arrangements were interactively laid out by a 3D artist using 3D assets from the TDW model library~\cite{gan2020threedworld}.  We will make this dataset publicly available.

\subsection{Objects}
To build physical house environments for the agent to navigate, we also populated the various floor plans with around 50 categories of objects. These include typical household items like lamps, toys, vases, pillows, printers, and laptops. Except for non-movable objects like walls, all objects respond to physics, so the embodied agent can interact with them.

\subsection{Embodiment}
We use the Magnebot as an embodied agent.  The Magnebot is equipped with an RGB-D camera and  also capable of returning image masks for object ID segmentation, semantic segmentation, normals and pixel flow.  For interactions with scene objects, the avatar has two articulated arms with 9-DOF ``magnet” end-effectors for picking up objects (see supplementary material for details). All motion and arm articulation actions are fully physics-driven.

The observation space in this challenge contains a first-person view RGB image, a depth map, and a semantic segmentation mask. The agent could use these observations to estimate the occupancy map (see below).  And it can also obtain an object list based on the semantic segmentation mask.

\subsection{Action space}
 We developed a high-level API with action commands designed to fit our task definitions, enabling an agent to move and manipulate objects. Our design choices were intended to abstract away the details of low-level control, which could otherwise form a barrier to exploring novel solutions for this kind of TAMP-style embodied AI challenge. 
 
 There are two types of actions we consider in this transport challenge: navigation and interactive actions. API functions for navigation include \emph{Move Forward By ($x$ m),  Rotate By ($\theta$ degrees), Rotate To (an object or target position)}. In our experiments, we make the actions discrete by setting $x =0.5 meter$, $\theta = \pm 15^\circ$.  \emph{Rotate to $+15^\circ$} and \emph{ Rotate to $-15^\circ$} means to rotate left or right by 15 degrees, respectively.  Interactive action functions include \emph{Go to Grasp, Drop} and \emph{Put In Container}. Taking as an example a simple task that involves the agent selecting an object to pick up based on the segmentation mask returned by sensors, going to grasping it, then moving to a container and dropping the object into the container, the action sequence we applied is \emph{Go to Grasp (object), Go To Grasp (container),} and \emph{Put In Container}. We use Inverse Kinematics to implement these interactive actions. The details can be found in the supplementary materials. 

Since our high-level action API is fully physics-driven, these features also pose new challenges that have not appeared in previous navigation environments, but indeed exist in the real physical world.  For example, the agent might fail to grasp one target object when we apply a \emph{Go to Grasp} action. The reason might be that its arm cannot reach the target object. In this case, the agent needs to move close to the object and make an attempt again.  We also observe that the objects/container held by the avatar can fall off if they hit a heavy obstacle during the navigation. The agent would then have to find and pick up these items again.  3433

\subsection{Action Status} 

When performing tasks, the simulation will also provide feedback on its status and the status of its current action, in order to decide what action to take next. Each API function returns an \emph{ActionStatus} value indicating the status of the agent -- whether it is performing an action, succeeded at an action, or failed to complete the action. There are 18 types of action status, including: whether a target object is too close or too far to reach, or is behind the agent; whether the agent succeeded or failed to pick up an object, or overshot the target; whether the agent collided with something heavy or is obstructed by an obstacle, etc. We can use these meta data to define a reward function for policy learning or priors for planning.

\subsection{Dataset Generation}

\noindent\textbf{Target Objects} The objects that the agent needs to  transport as part of the challenge task are known as ``target objects”.   These objects are located at ``target positions” within the environment, guaranteed to be reachable by the agent. The set of target objects to be transported are randomly placed across different rooms. In addition, there is a 25\% chance of spawning a container at a traversable position in each room. When a scene is initialized, the simulation loads a pre-calculated ``occupancy map”, i.e. a data file representing grid spaces occupied by objects. 
%This map spans the extent of the scene, and the size of the grid is equal to the radius of the agent. 
%Occupancy maps are generated by performing a ``spherecast” (basically a wide raycast) vertically down per grid position; any surfaces the spherecast hits are added to the map, along with their Y values (heights). 
%Grid positions at floor level are considered ``free”; regions of contiguous free positions are calculated and the largest region represents the set of potential target positions.

\noindent\textbf{Goal Positions} The locations within the environment where target objects must be deposited to complete the task are known as ``goal positions”. The surfaces that target objects are to be deposited on are known as ``goal position surfaces”.  Goal position surfaces can be on any of the following furniture objects: Sofa, Bench, Table, Coffee table, and Bed. For each task, we set one unique piece of furniture in the scene as the goal position (e.g. only one coffee table, or one sofa in the house) in the TDW-House dataset. %However to provide multiple options for goal position surfaces within the same room, often a sofa and bench will appear, plus a table and/or a coffee table.

\noindent\textbf{Robot Agent Initialization}
After the scene configuration is initialized, the robot agent is spawned at an open location that is free of other objects. 

\noindent\textbf{Task Definition}
We define the goal of each task with two components: 1) a set of objects and their counts, 2) the Goal Position. For instance, \emph{``vase:2, bowl:2, jug:1; bed"}  means that the agent must transport 2 vases, 2 toys, and 1 jug to the bed. There are five types of objects used as potential target objects to be transported. Each task requires transporting 6 to 8 objects.

\begin{figure}[!h]
    \centering
    \includegraphics[width=1.0\linewidth]{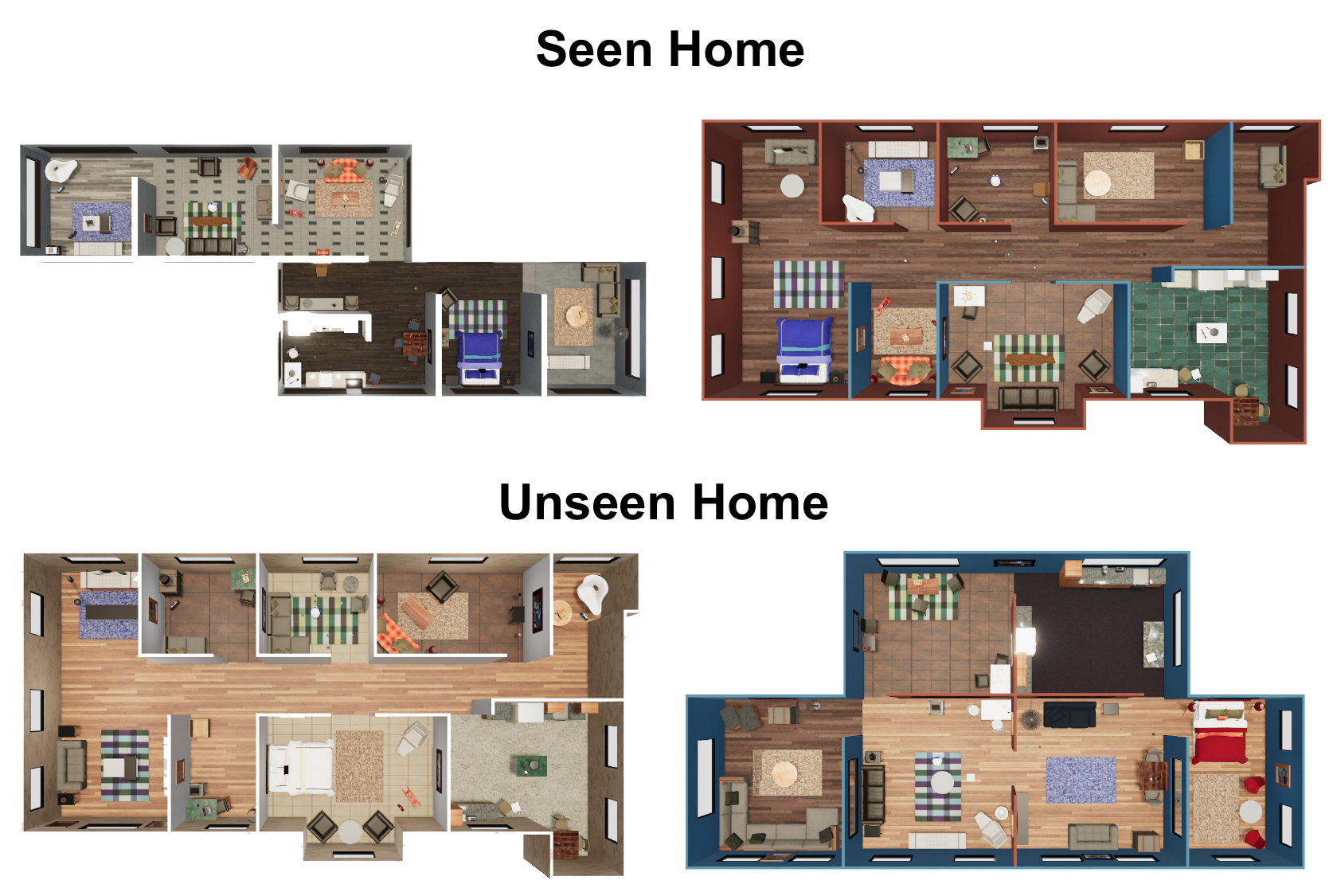}
    \caption{Example layouts of seen and unseen houses.}
    \vspace{-4mm}
    \label{fig:room}
\end{figure}
\section{Experiments}
In this section, we discuss our experimental setup, baselines, implementation details and evaluation results. 

\begin{figure*}[t]
    \centering
    \includegraphics[width=0.8\linewidth]{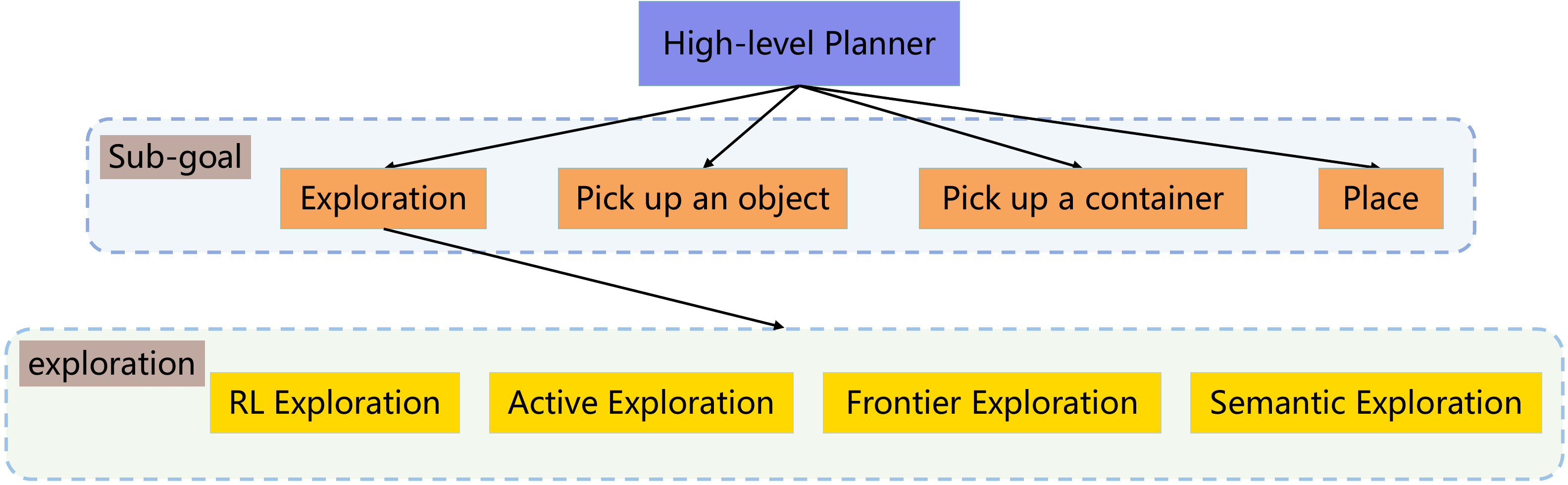}
    \vspace{-2mm}
    \caption{The flowchart of high-level and low-level planners.}
    \label{fig:framework}
\end{figure*}
\subsection{Experimental Setup}

\noindent\textbf{Setup.} We use 15 physically distinct houses for the experiments. Example house layouts can be found in Figure~\ref{fig:room}.  Since this challenge aims to test the agent's generalization abilities, we split them into 10 seen houses (training) and 5 unseen houses (test). We generated 100 tasks for each seen house and 20 tasks for each unseen house by randomly placing target objects and containers into different rooms as described above (for a total of 1000 training and 100 test tasks). We report the models' performance on the test set tasks.  

\noindent\textbf{Evaluation Metrics.}
The objective of this challenge is to transport the maximum number of objects as efficiently as possible. We use the \textbf{transport rate} as an evaluation metric. The transport rate is the fraction of the objects successfully transported to the desired position within a given interaction budget (defined as a maximum episode length in steps). For testing we set this maximum episode length to 1000 steps. 

\begin{figure*}[!ht]
    \centering
    \includegraphics[width=1.0\linewidth]{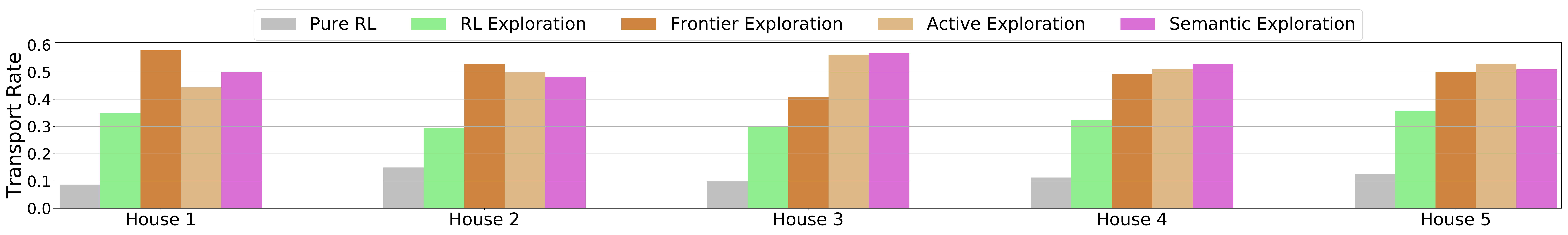}
    \caption{Comparisons of transport rates in each unseen room. }
    \label{fig:result}
\end{figure*} 
\subsection{Baseline Models}
We implemented several baseline agents using both learning- and planning-based algorithms. We found empirically that it was difficult for an end-to-end RL learning model (Pure RL baseline) to take observation data as input and directly search for an optimal policy over the agent’s action space, presumably because our task is very challenging (\eg long-horizon, partial observability, involving complex physical interaction, sparse rewards, \etc). Therefore, we implemented additional baselines using a hierarchical planning framework with different exploration strategies. 

As shown in Figure~\ref{fig:framework}, we define four types of sub-goals for this task: \emph{exploration, pick up a container, pick up an object,} and \emph{place}. We use a rule-based high-level planner to decide when to transition between sub-goals. For the \emph{exploration} sub-goal, the agent will either directly use the policy to navigate around the environment (RL Exploration baseline) or plan a shortest path from the current location to the waypoints (Active, Frontier and Semantic Exploration baselines). The agent can accumulate RGB images, depth maps, and segmentation masks to construct a 2D occupancy map and a semantic map during its exploration. When setting the sub-goal as \emph{pick up a container} or \emph{pick up an object},  the agent will plan a sequence of actions to move towards a container or a target object and pick it up. For the  \emph{place} sub-goal, the agent uses a planner to generate a deterministic local policy to go to the goal position and drop the target objects. We summarize the baseline models as follows: 
\begin{compactitem}

\item \textbf{Pure RL:} We train an end-to-end RL policy using PPO by maximizing the reward of finding objects (either target objects, containers, or the goal position), grasping objects, putting objects into a container, and dropping them onto the goal location. This model takes the inputs of observation and estimated occupancy map and directly outputs actions that the agent should execute in the environment. 

\item \textbf{RL Exploration:} We train a policy network that maximizes occupancy map coverage. The network yields an action to execute. 

\item \textbf{Frontier Exploration:} Similar to~\cite{yamauchi1997frontier}, the agent uses depth images and semantic segmentation masks to construct an occupancy map and a semantic map of the scene. The agent randomly samples a waypoint from an unexplored area as a sub-goal for exploration. It marks the location of target objects, containers, and the goal position on its semantic map.  %The planner can then produce a sequence of actions to reach the sub-goals and finally transport the objects to the goal locations.

\item \textbf{Active Exploration:}  We adapted a baseline from~\cite{chaplot2020learning}. Instead of randomly choosing an unexplored position in the occupancy map as a goal location to explore (as in the Frontier baseline), the Active baseline learns a goal-agnostic policy by maximizing map coverage (using neural SLAM). The output is a waypoint position. The local exploration planner uses this waypoint as a sub-goal.  

\item \textbf{Semantic Exploration:}  We adapted a baseline from~\cite{chaplot2020object}. The Semantic baseline uses neural SLAM to learn an object/goal-oriented ``semantic" exploration policy. The reward function encourages the agent to find more target objects, containers, and the goal position. 
\end{compactitem}

\subsection{Implementation Details}
For all the baselines, the input observation data are 256 $\times$ 256 size RGB images, depth maps, and semantic segmentation masks.  The action spaces include 3 navigation actions (\ie move forward, rotate left, and rotate right) and 3 interactive actions (\ie go to grasp, put in container, and drop). Below we provide the details of the occupancy map, semantic map, policy training, high-level and low-level planners. 

\noindent\textbf{Occupancy Map.}
We represent the global top-down occupancy map as $ O_g \in R^{2 \times N \times N}$, where N$\times$N denotes the
map size and each element in this occupancy map corresponds to a region of size 0.25m$\times$0.25m in the physical world.  We can recover a local occupancy map $ O_l \in R^{
2 \times K\times K}$ from egocentric depth map, which comprises a local area in front of the camera.  The two channels in each cell represent the probability scores of the cell being occupied and explored, respectively.  We considered a cell to be occupied if there is an obstacle and a cell to be explored if the agent already knows whether it is occupied or not. The occupancy map is initialized with all zeros and the agent always starts at the center of the map at the beginning of each episode.

\noindent\textbf{Semantic Map.} Similar to the occupancy map, we can also construct a global top-down semantic map $ S \in R^{3 \times N \times N}$ using both depth images and semantic segmentation masks. The value in each channel indicates if a cell contains target  objects, containers or goal position. Similar to the occupancy map, we also initialized the semantic map with all zeros and set the agent starting position as the map center. The agent can gradually update the semantic map when they find objects during the navigation exploration.

\noindent\textbf{Policy learning.} For all policy learning, we use a four-layer CNN network and train for 10 million frames using Proximal Policy Optimization~\cite{schulman2017proximal}. In the Pure RL, RL Exploration, and Semantic Exploration baselines, we take the depth image, estimated occupancy map, and semantic map as input. In the Active Exploration baseline, we take the depth image and estimated occupancy map as inputs.  In the Pure RL and RL Exploration baselines, the model directly outputs an action. For Active and Semantic Exploration baselines, the models predict waypoints as sub-goals. We use the same reward function for the Pure RL and Active Exploration baselines by encouraging the agent to find new objects, grasp target objects, put objects into a container, and drop objects onto the goal position. In the RL Exploration and Active Exploration baselines, we use a goal-agnostic reward function by encouraging the agent to improve the map coverage.

\begin{figure*}[!h]
    \centering
    \includegraphics[width=1.0\linewidth]{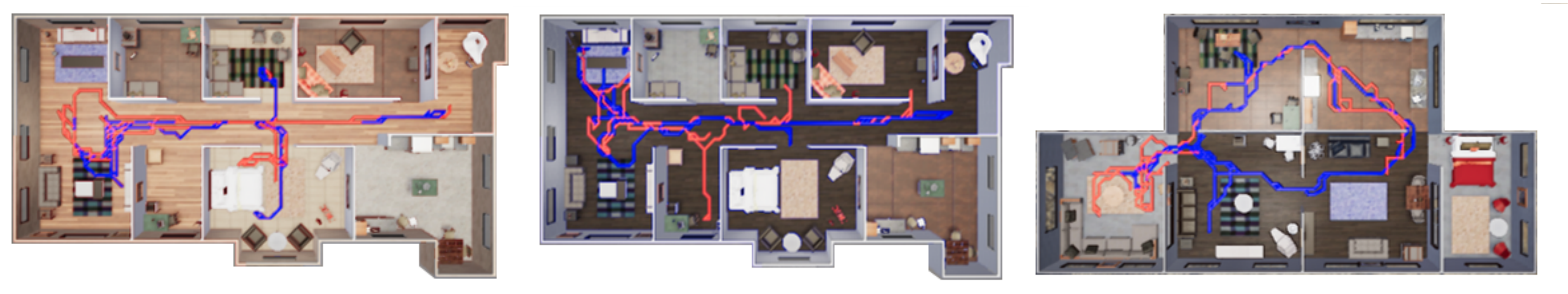}\vspace{-2mm}
    \caption{Navigation trajectories of frontier exploration (blue) against those of RL exploration (red). The frontier exploration finds more efficient routes to transport more objects.}
\vspace{-4mm}
    \label{fig:trajectory}
\end{figure*}

\noindent\textbf{High-level planner.} We define a rule-based high-level planner to decide the sub-goal. The agent takes \emph{exploration} as a sub-goal by default.  At the beginning, the sub-goal can switch from \emph{exploration} to \emph{picking up a container} once a container has been found. After a container is found or 20\% of the interaction budget has been used up, the agent can switch the sub-goal from \emph{exploration} to \emph{picking up an object} if they find a target object. After 90\% of the interaction steps, the agents will switch their sub-goal to \emph{place}.  If there is no container found, the agent will switch their sub-goal from \emph{picking up an object} to \emph{place} if they hold two objects. 

\noindent\textbf{Low-level planner.} We use a A-Star based planner to find the shortest path from the current location to the sub-goal location and execute the high-level interactive actions for achieving the sub-goal. Every interaction steps, we update the map and re-plan the path to the sub-goal.

\begin{table}[t]
    \centering
    \caption{Comparison of transport rate over 5 unseen rooms.}
    
    \begin{tabular}{l|c}
   \hline
 Method & Transport Rate  \\
 \hline
 Pure RL    &  0.12$_{\pm0.01}$\\
 RL Exploration & 0.33$_{\pm0.02}$ \\
 
  Frontier Exploration &  0.50$_{\pm0.03}$\\
Active Exploration  &  0.51$_{\pm0.01}$\\Semantic Exploration  & 0.52$_{\pm0.01}$   \\

    \end{tabular}
    \label{tab:main}
\end{table}
\subsection{Result Analysis}
In this section, we first present the overall benchmark results, and then perform in-depth analysis of the model design. We hope that our experimental findings will provide useful benchmarks for physically realistic Embodied AI.

\noindent\textbf{Overall Results.} We first plot the histogram of transport rate for all five unseen houses in Figure~\ref{fig:result}. We calculate the results by averaging over all 20 testing tasks in each house. We also summarize the overall quantitative evaluation results of all baseline models measured by average transport rates of 5 unseen rooms in Table~\ref{tab:main}. The best model achieves an average of 0.52 transport rates under 1000 episodes.  We have two key observations: 1) There are no agents that can successfully transport all the target objects to the goal locations. This means our proposed task is very challenging and could be used as a benchmark to track the progress of embodied AI in physically realistic scenes.  2) The Pure RL baseline performs poorly. We believe this reflects the complexity of physical interaction and the large exploration search space of our benchmark. Compared to the previous point-goal navigation and semantic navigation tasks, where the agent only needs to navigate to specific coordinates or objects in the scene, the ThreeDWorld Transport challenge requires agents to move and change the objects' physical state in the environment(\ie task-and-motion planning), which the end-to-end models might fall short on. 

\noindent\textbf{Exploration policy.} The results also indicate that combining RL exploration policy into a hierarchical planning framework could significantly improve results over a purely RL-based approach (better performance of all Exploration baselines over Pure RL). However, predicting a way-point (Frontier, Active, and Semantic Exploration) produced better results than predicting an action (RL Exploration). We speculate that RL Exploration can find different positions, but that it is hard for this model to learn complex physical interactions. In practice, we observed that this agent always failed to grasp objects, and frequently collided with the obstacles, resulting in a low transport rate. These results support the idea that our benchmark can enable novel algorithms to combine learning-based and planning-based methods to solve complex tasks in physically realistic environments. 

\noindent\textbf{Neural-based Exploration.} 
We also found that learning to generate a waypoint for exploration (in the Active and Semantic Exploration baselines) did not show clear advantages over the model without learning (Frontier Exploration). The reasons might be two-fold: 1) the layout of our test scenes are quite different from those of the training scenes. 2) The occupancy and semantic maps used for learning method do not convey much physics information. We believe future work that explicitly incorporates physics information might help improve the generalization abilities of learning-based methods. 

\noindent\textbf{Qualitative Results.} We further visualize the trajectory of different agents in Figure~\ref{fig:trajectory}. We can observe that the planning-based agent using waypoints as sub-goals can always find the shortest path to grasp objects and transport them to the goal locations. The RL agent is good at exploration, but might frequently hit obstacles, thus fail to transport objects efficiently.  We will also provide demo videos in the supplementary materials.

% \subsection{Human Evaluation}
% We obtained a human evaluation of 100 randomly sampled directives from the unseen test fold. The experiment
% involved 5 participants who completed 20 tasks each using a keyboard-and-mouse interface. Before the experiment, the participants were allowed to familiarize themselves with TDW. The action-space and task restrictions were identical to that of the baseline models. Overall,
% the participants could easily achieve a tansport rate over 90\%.  This indicates that current AI model are still fall short on such task-and-motion planing task that require advanced cognitive skill.

\section{Conclusions and Future Work}
We introduce a visually-guided task-and-motion planning  benchmark that we call the ThreeDWorld Transport Challenge. Our preliminary experimental results indicate that the proposed challenge can assess AI agents' abilities to rearrange multiple objects in a physically realistic environment. We believe our benchmark will also remove the barrier to entering the field of Task and Motion Planning (TAMP), allowing more people to study such a challenging TAMP-style embodied AI task in the face of realistic physical constraints. We plan to include deformable or soft body objects in future versions of this challenge. 

{\small
\bibliographystyle{ieee_fullname}
\bibliography{egbib}
}
\newpage

\appendix

%%%%%%%%% TITLE

\section{Magnebot Details} 
At a low level, the Magnebot is driven by robotics commands such as \textbf{set\_ revolute\_target}, which will turn a revolute drive. The high-level API combines the low-level commands into "actions", such as \textbf{grasp(target object)} or \textbf{move\_by(distance)}. 

To facilitate reaching and grasping we utilize an inverse kinematics (IK) solver to reach the desired target object or position. 

The Magnebot end-effectors achieve 9 degrees of freedom from a 3-DOF spherical joint at its ``shoulder", a 1-DOF revolute joint at its ``elbow", and a 3-DOF spherical joint at its "wrist"; the two additional DOF come from the ability to move the entire torso structure up and down the central column, as well as rotate the torso left/right about the column axis.

The Magnebot moves by turning its wheels, each of which is a revolute drive. When turning, the left wheels will turn one way and the right wheels in the opposite way, allowing the Magnebot to turn in place.

\section{Implementation Details of High-Level API}
The ThreeDWorld Transport Challenge is built on top of the ThreeDWorld (TDW), a high-fidelity platform for interactive physical simulation.

\subsection{TDW system overview.} 
A TDW simulation consists of two basic components: (i) the \textbf{Build} or runtime simulation engine; and (ii) the \textbf{Controller}, an external Python interface to communicate with the build. Running a simulation follows a cycle in which:
1) The controller sends \textbf{commands} to the build; 2) The build executes those commands and sends \textbf{simulation output data} back to the controller.

\subsection{High-level API.} 

TDW's Python API contains over 200 commands covering a broad range of use case scenarios. We can use this low-level ``atomic" API as building-blocks to create controller programs to run simulations. For the ThreeDWorld Transport Challenge we created a new set of high-level commands built on top of TDW's existing API, significantly reducing the number of commands required to perform complex actions such as grasping objects and placing them into containers. The Transport Challenge follows the OpenGym API convention. Each executed action will return new observation and scene state data as well as a status value indicating whether the action succeeded or failed. The Transport Challenge action space includes:

\begin{compactitem}

% By default, the target is relative to the agent's position and rotation.

% Once placed in a container, an object will be permanently attached to the container; it cannot be removed again.

\item \textbf{move\_forward()} Move forward by a given distance. \\

\item \textbf{rotate\_left()} Rotate left by a given angle. \\

\item \textbf{rotate\_right()} Rotate right by a given angle. \\

\item\textbf{go\_to\_grasp():} Move to a target object or container and bend the arm using an inverse kinematics (IK) solver to grasp the object.\\

\item \textbf{drop()}  Drop an object held by the arm.\\

\item \textbf{put\_in\_container()} Bend the agent's arms using an IK solver in order to put an object held by one magnet into a container held by another magnet.
%     \item The agent will pick up the container again.

\end{compactitem}

\section{Full List of Action Status Values} The high-level API functions in this simulation platform return ActionStatus values indicating whether the action succeeded or failed (and if so, why). Some of these values are not used in the Transport Challenge API. We can use this meta data to define reward functions. The full set of ActionStatus values includes:

\begin{compactitem}
		\item   \emph{ongoing}: The action is ongoing.

		\item   \emph{success}: The action was successful.

		\item   \emph{failed\_to\_move}: Tried to move to a target position or object but failed.

		\item   \emph{failed\_to\_turn}: Tried to turn but failed to align with the target angle, position, or object.

		\item   \emph{cannot\_reach}: Didn't try to reach for the target position because it can't.

		\item   \emph{failed\_to\_reach}: Tried to reach for the target but failed; the magnet isn't close to the target.

		\item   \emph{failed\_to\_grasp}: Tried to grasp the object and failed.

		\item   \emph{not\_holding}: Didn't try to drop the object(s) because it isn't holding them.

		\item   \emph{clamped\_camera\_rotation}: Rotated the camera but at least one angle of rotation was clamped.

		\item   \emph{failed\_to\_bend}: Tried to bend its arm but failed to bend it all the way.

		\item   \emph{collision}: Tried to move or turn but failed because it collided with the environment (such as a wall) or a large object.

		\item   \emph{tipping}: Tried to move or turn but failed because it started to tip over.

		\item   \emph{not\_in}: Tried and failed to put the object in the container.

		\item   \emph{still\_in}: Tried and failed to pour all objects out of the container.

\end{compactitem}

\section{SceneState data} 

In addition to the ActionStatus values, this simulation platform also provides per-action SceneState data per-frame data that an agent can use to decide what action to take next. This data includes:
\begin{compactitem}
	\item Visual data – rendered image, object color segmentation and depth values as numpy arrays
	\item The projection matrix of the agent's camera as a numpy array
	\item The camera matrix of the agent's camera as a numpy array
	\item The positions and rotations of each object in the scene
	\item The position and rotation of the Magnebot
	\item The position and angle of each of the Magnebot's joints
	\item Which objects are currently held by each of the Magnebot's magnets
\end{compactitem}

% \section{Challenge API Code Examples}
% In the following code examples we show the basic steps required to instantiate a scene, grasp objects and place them into containers.\\ 

% \noindent\textbf{Scene initialization example.}This is a simple example of how to initialize an interior environment populated by furniture, objects, and an agent.

% \begin{verbatim}
% from sticky_mitten_agent import StickyMittenagentController

% if __name__ == "__main__":
%     # Instantiate the controller.
%     c = StickyMittenagentController(launch_build=False)
%     # Initialize the scene. Populate it with objects. Spawn the agent in a room.
%     c.init_scene(scene="2a", layout=1, room=1)
% \end{verbatim}

\end{document}